\documentclass[runningheads]{llncs}

\usepackage[T1]{fontenc}
\usepackage{graphicx}
\usepackage{float}
\usepackage{multirow}
\usepackage{booktabs}
\usepackage{placeins}
\usepackage{amsmath,amssymb,amsfonts}
\usepackage{xcolor}

\begin{document} \title{Dual-Selective Network for Domain-Incremental Change Detection} \author{Yuzhi He, Junxi Huang, Haorui Wu, Jiahui Qu$^\dagger$} \authorrunning{He et al.} \institute{ Xidian University, Xi'an, China\\ \email{\{23012100059,24012100067,23012100073\}@stu.xidian.edu.cn, jhqu@xidian.edu.cn} \quad $^\dagger$Corresponding author }

\maketitle            

\begin{abstract}
Domain-incremental change detection (DICD) continuously adapts models to new geographic domains while preserving prior knowledge. However, a structural mismatch exists: the label space remains fixed while domain characteristics vary drastically. Consequently, incremental models struggle to maintain stable spatial change representations across domains. Existing strategies---such as replay-based or regularization-based methods---often fail to scale to long domain sequences, leading to knowledge degradation or increased computational cost. We propose \textbf{Dual-Selective Incremental Network} (DSINet), a unified framework built on visual state space models. DSINet leverages Mamba's input-dependent selective mechanism through a selective spatial state unit (S$^3$U). This unit preserves stable spatial change structures while filtering domain-specific variations during feature propagation. As a result, spatial representations remain stable across domains, preventing the accumulation of feature confusion over incremental steps. Additionally, we employ a concentration-balanced distillation (CBD) strategy to stabilize knowledge transfer across domains. It balances hardness and confidence concentration effects during incremental updates. This ensures reliable probability mass allocation and prevents over-smoothing or mode collapse during distillation. Together, these mechanisms maintain stable learning dynamics throughout incremental stages. Experimental results demonstrate that DSINet mitigates knowledge degradation across long domain sequences while maintaining the linear computational efficiency of state space models.
\end{abstract}
\section{Introduction}

\begin{figure}[htbp]
  \centering
  \includegraphics[width=1.0\textwidth]{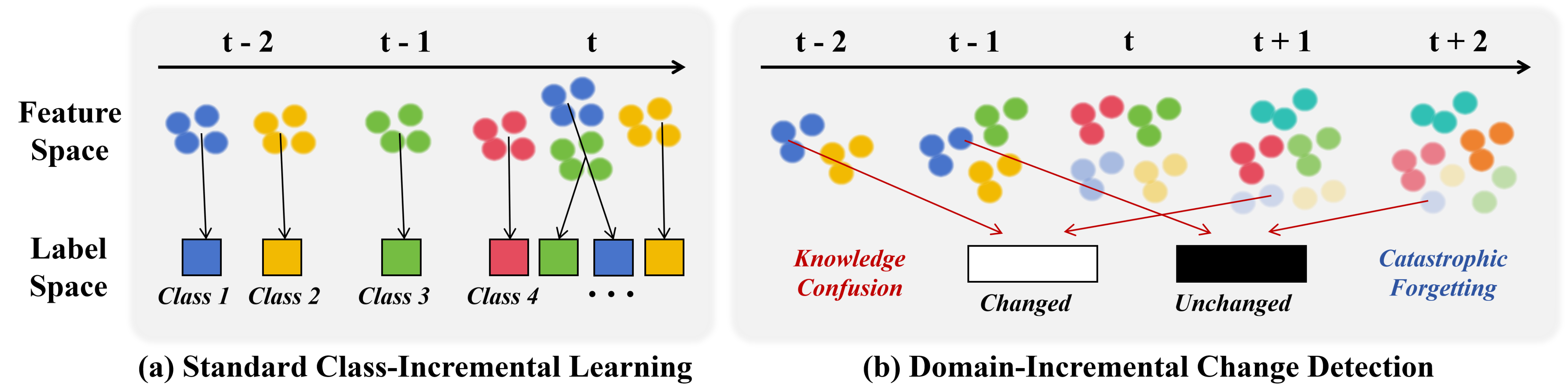}
  \caption{Comparison between standard incremental learning and DICD (a) Standard incremental learning: both feature and label spaces expand with new classes. (b) DICD: fixed binary label space with shifting feature domains.}
  \label{fig1}
\end{figure}

% 第一部分：引出研究背景与本文聚焦的核心问题 (Domain-Incremental Change Detection)
Change detection (CD) monitors earth surface dynamics and remains fundamental to remote sensing \cite{Cheng2024Review,Chen2024ChangeMamba}. In practice, CD models must continuously integrate data from newly observed geographic regions. This requirement naturally motivates continual learning. Task-incremental learning (TIL) relies on explicit task identifiers, while class-incremental learning (CIL) expands the categorical label space \cite{vandeVen2022Three}. In contrast, domain-incremental learning (DIL) maintains a consistent learning objective while accommodating variations in input data distributions \cite{vandeVen2022Three,Huang2024Domain}. For practical CD applications, domain-incremental change detection (DICD) is particularly relevant \cite{Weng2024MDINet}. As illustrated in Fig.~\ref{fig1}, DICD maintains a fixed binary label space (changed vs. unchanged), while the input feature distribution shifts across different geographic domains. However, this setting introduces a structural mismatch. The discrepancy between a constant label space and shifting feature distributions causes spatial representation confusion---newly encountered domain-specific variations overwrite the domain-agnostic structural knowledge essential for fundamental change identification.

%% ---------------- 后续段落写作规划 ----------------
%% 1. 介绍目前做这个问题（或者相关CD增量学习）的主流研究进展。
%% 2. 分析这些主流方法分别存在的缺点和弊端（例如：对旧域知识的灾难性遗忘、无法有效处理域间特征偏移等）。
%% 3. 借由上述弊端，顺理成章地引出本文提出的方法及其核心优势。
%% --------------------------------------------------

% 第二部分：现有主流增量学习方法的局限性 (Replay & Regularization / KD)
Existing continual learning frameworks primarily mitigate catastrophic forgetting through data replay or regularization \cite{Wang2023Survey}. While effective for short domain sequences, they encounter stability challenges when scaled to longer trajectories. Replay-based methods preserve prior knowledge by maintaining a memory buffer of historical samples \cite{Lee2025ERPASS}. However, under strict memory constraints, the representation of earlier domains becomes progressively diluted as new domains are encountered. This data sparsity restricts spatial diversity, causing the model to overfit on a limited subset rather than generalizing past structural knowledge. Alternatively, regularization-based methods, particularly knowledge distillation (KD) \cite{Himeur2025Survey}, preserve prior knowledge by constraining network updates. Yet, when spatial features remain entangled and probability mass is rigidly matched via standard Kullback-Leibler Divergence \cite{Wang2025ABKD}, the model suffers from representation confusion and distribution over-smoothing. Over extended sequences, these step-wise inaccuracies accumulate, degrading the historical teacher model and causing it to transfer unreliable structural priors to subsequent tasks. Consequently, scaling DICD to long sequences requires explicitly stabilizing both spatial feature propagation and knowledge transfer at every incremental step to mitigate error accumulation.

%% ---------------- 后续段落写作规划 ----------------
%% 1. 引出本文提出的具体方法（结合了 Mamba 架构和特定的增量学习机制）。
%% 2. 简述本文方法是如何解决上述提到的“空间特征纠缠”和“误差累积”问题的。
%% 3. 列出本文的核心贡献 (Contributions)。
%% --------------------------------------------------

% 第三部分：引出本文方法 (总-分结构：总述框架 -> 空间特征选择 -> 优化目标选择)
To address these challenges, we design the DSINet, a unified teacher-student framework built upon SSMs \cite{Zhao2024RSMamba,Chen2024ChangeMamba}. By establishing rigorous step-wise stability across both spatial feature propagation and probabilistic knowledge transfer, DSINet effectively mitigates representation confusion and error accumulation in long domain sequences. 

At the spatial level, we design the S$^3$U. Instead of relying on rigid parameter isolation, S$^3$U leverages the input-dependent selective mechanism of SSMs to dynamically assign update step sizes. This active selection inherently preserves domain-agnostic geographic structures while filtering out domain-specific environmental noise. Consequently, S$^3$U stabilizes spatial representations and prevents the feature-level confusion that typically triggers catastrophic forgetting.

At the optimization level, we introduce CBD to ensure stable knowledge transfer. To overcome the distribution matching dilemma in traditional distillation---which often leads to over-smoothing or background mode collapse---CBD reformulates probability matching as a selective allocation process via the $\alpha$-$\beta$ divergence framework \cite{Wang2025ABKD}. By dynamically balancing hardness-concentration (regions with high teacher-student discrepancies) and confidence-concentration (structurally certain predictions), CBD acts as a probabilistic filter. Together, these dual-selective mechanisms enable the student network to inherit reliable structural priors without degradation, allowing DSINet to scale robustly across continuous domains while maintaining the $\mathcal{O}(N)$ linear computational efficiency of its SSM backbone. To summarize, the contributions of this work are as follows:
% 第四部分：本文的核心贡献 (Main Contributions)
\begin{itemize}
    \item We propose \textbf{DSINet}, a dual-selective framework designed to address the challenges of representation confusion and distillation instability in long domain sequences. The framework explicitly maintains step-wise stability to mitigate severe error accumulation across prolonged domain shifts.
    \item We design the \textbf{S$^3$U} to overcome spatial knowledge confusion. By leveraging the data-dependent selection mechanism of SSMs, it preserves universal geographic structures while filtering domain-specific variations, ensuring more stable feature propagation than static architectures.
    \item We introduce \textbf{CBD} to stabilize knowledge transfer. By treating probability matching as a selective allocation process, CBD balances hardness and confidence concentration to prevent distribution over-smoothing and mode collapse. Extensive experiments demonstrate that DSINet effectively mitigates knowledge degradation without the prohibitive costs of large-scale data replay.
\end{itemize}

\section{Methodology}

\begin{figure}[h]
\centering
\includegraphics[width=1.0\textwidth]{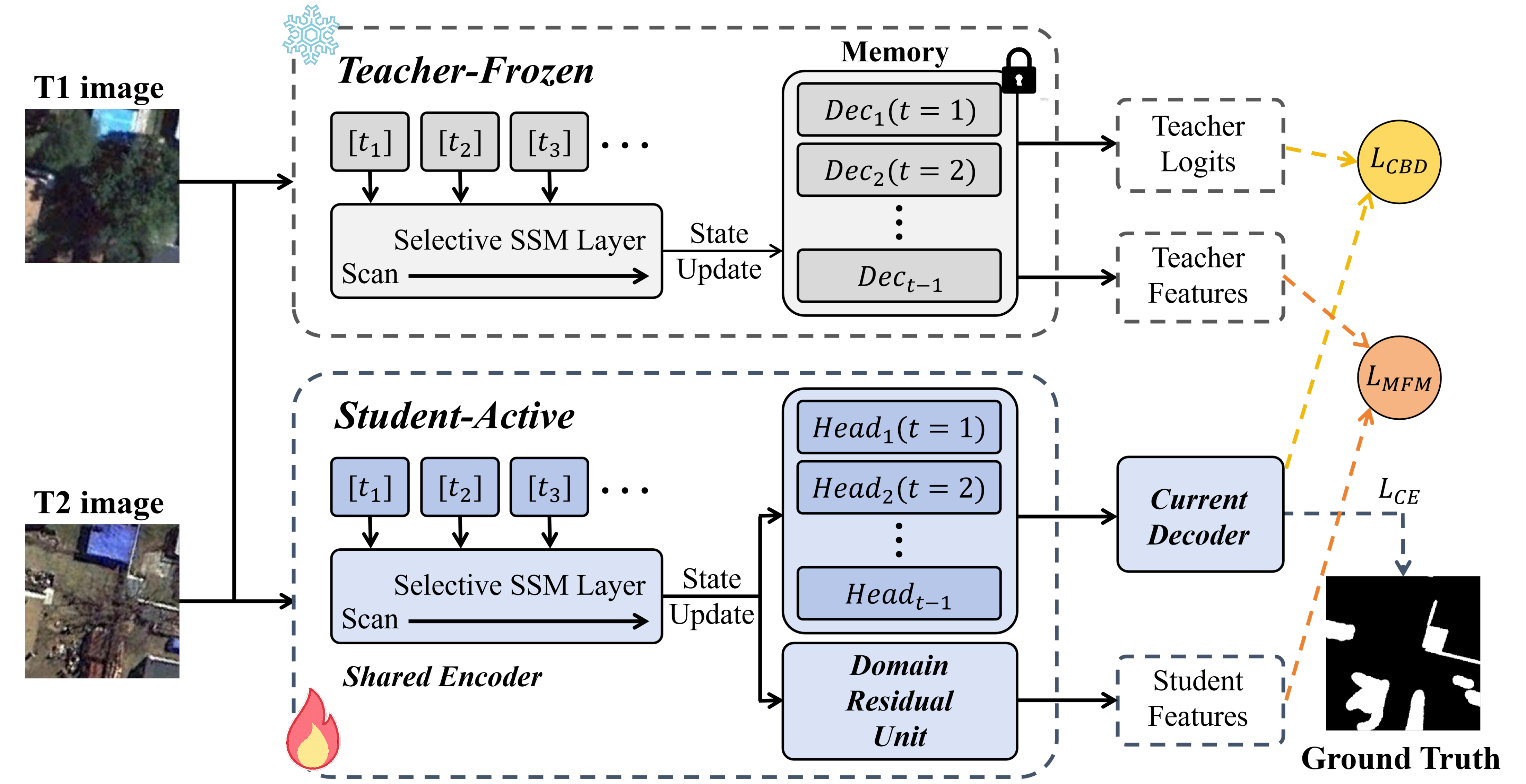} 
\caption{Overview of the proposed DSINet. Designed for long domain sequences, DSINet maintains step-wise stability through two selection mechanisms. }
\label{fig:overview}
\end{figure}

\subsection{Problem Formulation and Dual-Selective Architecture}
The objective of DICD is to enable a model to sequentially learn mapping functions across continuous geographical domains $\mathcal{D}_1, \mathcal{D}_2, \dots, \mathcal{D}_T$ without catastrophic forgetting. At each incremental step $t$, the model receives bi-temporal images $\mathcal{X}_t = \{x_1, x_2\}_t$ from a new domain $\mathcal{D}_t$ and outputs a binary prediction map $y_t \in \{0, 1\}^{H \times W}$. In the context of long domain sequences, standard incremental updates inevitably lead to severe error accumulation. Therefore, maintaining rigorous step-wise stability is critical to prevent historical representations from being overwritten by continuous environmental shifts.

To resolve spatial knowledge confusion and distribution mismatch in long-sequence DICD, we introduce \textbf{DSINet}, whose overall framework is illustrated in Fig.~\ref{fig:overview}. The network adopts a teacher-student Siamese structure consisting of a shared SSM-based encoder $F$ and domain-specific decoders $G_t$. The forward pass at step $t$ is formulated as:
\begin{equation}
    M_t(\mathcal{X}_t; W_s, W_t) = G_t(F(\mathcal{X}_t; W_s, \phi_t)),
\end{equation}
where $W_s$ represents the domain-shared parameters, and $W_t = \{\phi_t, G_t\}$ denotes the domain-specific parameters for step $t$. Supported by the $\mathcal{O}(N)$ linear complexity of the 2D cross-scan mechanism, this architecture renders high-resolution dual-model processing computationally efficient. 

Crucially, rather than relying on static regularization, DSINet ensures step-wise stability through two orthogonal active selection mechanisms: the S$^3$U for dynamic spatial feature disentanglement (detailed in Section \ref{sec:s3u}), and CBD for selective probability mass allocation during knowledge transfer (detailed in Section \ref{sec:cbd}).

\subsection{Representational Level: Selective Spatial State Unit} \label{sec:s3u}
\begin{figure}[htbp]
\centering
\includegraphics[width=1.0\textwidth]{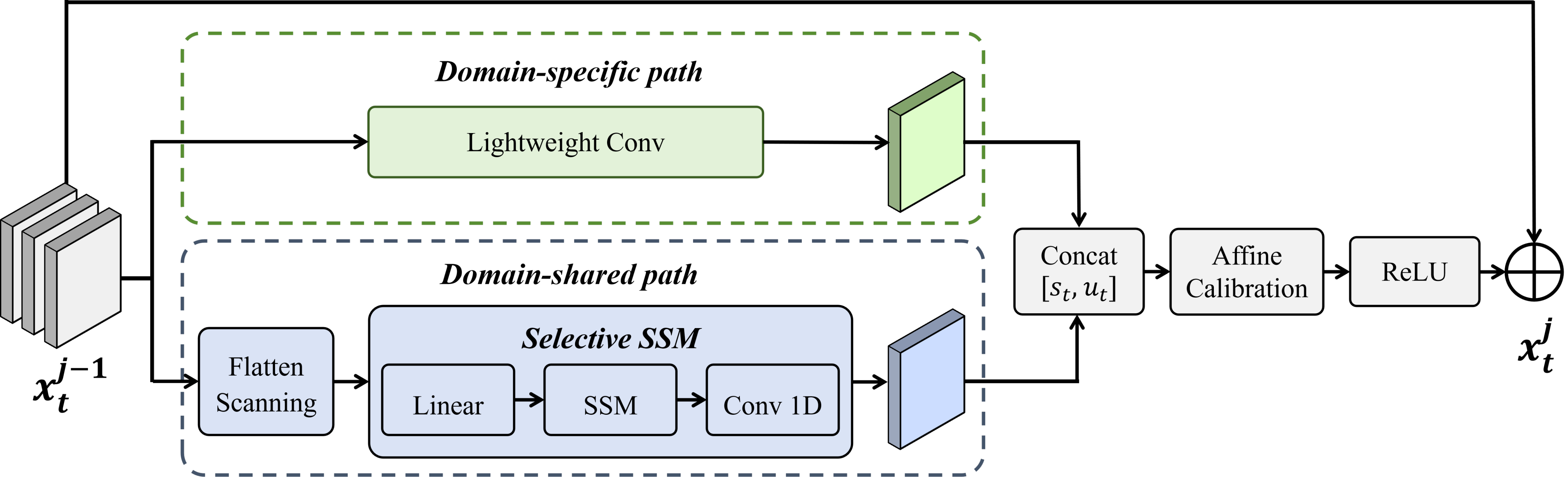} 
\caption{Structure of the S$^3$U. The input feature map is dynamically disentangled into two parallel pathways. The domain-shared pathway utilizes the 2D cross-scan mechanism of SSMs to capture global domain-agnostic structures. Simultaneously, the domain-specific pathway extracts localized variations using a lightweight convolution. The concatenated features are calibrated via channel-wise affine transformations, projected back to the original dimension, and fused via a residual connection.}
\label{fig:s3u}
\end{figure}

In incremental CD, maintaining a strictly binary label space while feature distributions shift across continuous domains naturally leads to optimization divergence. To address this, the network must explicitly disentangle domain-agnostic structural representations from domain-specific spectral variations. Unlike traditional domain residual units (DRUs) that rely on static parameter isolation, we design the S$^3$U to dynamically route and filter features based on their spatial context as shown in Fig.~\ref{fig:s3u}. 

\subsubsection{Selective State Space Mechanism.}
The foundational SSM maps a 1D input sequence $x(t) \in \mathbb{R}$ to an output $y(t) \in \mathbb{R}$ through a latent state $h(t) \in \mathbb{R}^N$. This process is formulated as a linear continuous-time ordinary differential equation (ODE):
\begin{equation}
    h'(t) = \mathbf{A}h(t) + \mathbf{B}x(t), \quad y(t) = \mathbf{C}h(t).
\end{equation}

To integrate this into a deep learning framework, the continuous parameters ($\mathbf{A}, \mathbf{B}$) are discretized using a timescale parameter $\mathbf{\Delta}$ via the zero-order hold rule. Unlike standard SSMs, the visual Mamba architecture introduces a data-dependent selection mechanism \cite{Zhao2024RSMamba,Chen2024ChangeMamba}. Specifically, the matrices $\mathbf{B}$, $\mathbf{C}$, and crucially the timescale $\mathbf{\Delta}$, are parameterized as functions of the input. This selectivity enables the network to dynamically filter information based on the current spatial context rather than applying a uniform transformation.

\subsubsection{Dynamic Feature Disentanglement.}
We embed this selective capability into our parameter decomposition strategy. Let $x_t^{j-1} \in \mathbb{R}^{B \times C \times H \times W}$ denote the input feature map at the $(j-1)$-th module. The S$^3$U decomposes the representation into two parallel pathways:

The domain-shared pathway extracts generalized change representations using the Selective SSM, denoted as $W_{SSM}(\cdot)$. By flattening the spatial dimensions into a 1D sequence via a 2D cross-scan mechanism, the model assigns larger step sizes $\mathbf{\Delta}$ to low-frequency, domain-agnostic structures (e.g., universal building footprints), actively updating the hidden state $h$. Conversely, it assigns smaller step sizes to domain-specific environmental noise, effectively treating them as transient variations to be ignored.

The domain-specific pathway employs a lightweight convolution $\mathcal{C}_t^{j-1}(\cdot)$, initialized exclusively for domain $t$, to capture localized regional attributes. The domain-agnostic structures and domain-specific variations are then concatenated along the channel dimension and calibrated using learnable channel-wise scaling ($\gamma^j \in \mathbb{R}^{2C}$) and shifting ($\delta^j \in \mathbb{R}^{2C}$) parameters:
\begin{equation}
    D_{concat} = \gamma^j \odot \left[ \mathcal{C}_t^{j-1}(x_t^{j-1}), W_{SSM}(x_t^{j-1}) \right] + \delta^j,
\end{equation}
where $[\cdot]$ denotes channel concatenation and $\odot$ represents the Hadamard product. To align the dimensionalities for the subsequent residual connection, the calibrated $2C$-channel feature is reduced back to $C$ channels via a linear projection layer $\mathcal{P}(\cdot)$ (e.g., a $1\times1$ convolution) and followed by a non-linear activation:
\begin{equation}
    x_t^j = x_t^{j-1} + \text{ReLU}(\mathcal{P}(D_{concat})).
\end{equation}
Through this structurally rigorous mechanism, S$^3$U ensures that the shared parameters $W_s$ continuously accumulate generalized geographical features, while the specific parameters $\phi_t$ absorb shifting distributions, fundamentally preventing the feature-level overlap that causes catastrophic forgetting.

\subsection{Optimization Level: Concentration-Balanced Distillation} \label{sec:cbd}
Preserving historical knowledge requires effective distillation between a frozen teacher model $M_{t-1}$ and an active student $M_t$. However, in continuous CD tasks, standard distillation is hindered by a distribution matching dilemma.

\subsubsection{The Distribution Matching Dilemma.}
Standard knowledge distillation typically minimizes the forward Kullback-Leibler divergence (FKLD). From a gradient optimization perspective, FKLD allocates probability mass evenly, exhibiting a weak concentration effect. This forces the student distribution to become overly smooth, diluting the sharp structural boundaries transferred from the teacher. Conversely, utilizing reverse KLD (RKLD) prioritizes modes where the student is already confident, often driving the model into mode collapse.\cite{Wang2025ABKD} This causes the student to ignore the broader, soft-label structural information captured by the teacher.

\subsubsection{$\alpha-\beta$ Divergence for Selective Allocation.}
To resolve this dilemma, we introduce the CBD strategy, formulated upon the $\alpha-\beta$ divergence framework \cite{Wang2025ABKD}. For a teacher distribution $p$ and a student distribution $q_\theta$, the divergence is defined as:
\begin{equation}
\begin{gathered}
\mathbb{D}_{AB}^{(\alpha,\beta)}(p \parallel q_\theta) \triangleq \\[4pt]
\quad -\frac{1}{\alpha\beta} \sum_k \left[ p(k)^\alpha q_\theta(k)^\beta 
- \frac{\alpha}{\alpha+\beta} p(k)^{\alpha+\beta} 
- \frac{\beta}{\alpha+\beta} q_\theta(k)^{\alpha+\beta} \right].
\end{gathered}
\end{equation}
where $\alpha, \beta \in \mathbb{R}$ and $\alpha+\beta \neq 0$. The singularities at $\alpha=0$ or $\beta=0$ are rigorously resolved via continuous extension using L'Hôpital's rule. This generalized family smoothly recovers FKLD as $(\alpha \to 1, \beta \to 0)$ and RKLD as $(\alpha \to 0, \beta \to 1)$.

Crucially, these hyperparameters decouple two competing optimization forces. The parameter $\alpha$ controls \textit{hardness-concentration}: decreasing $\alpha$ amplifies the penalty on regions with large teacher-student discrepancies. Simultaneously, $\beta$ controls \textit{confidence-concentration}: increasing $\beta$ focuses gradient updates on predictions where the student is highly confident. By selecting intermediate values (e.g., $\alpha, \beta \in (0,1)$), CBD selectively allocates probability mass. It penalizes structural errors without over-smoothing, while preventing the student from collapsing into dominant background modes.

\subsubsection{Comprehensive Distillation Objective.}
While CBD stabilizes logit-level semantic transfer, spatial consistency across incremental stages must also be strictly enforced. To complement CBD, we integrate a multi-layer feature map (MFM) loss to anchor intermediate spatial representations. It computes the normalized mean squared error (MSE) between the teacher and student features:
\begin{equation}
    \mathcal{L}_{MFM} = \sum_{l \in L} \frac{1}{C_l H_l W_l} \left\| F^l_{t-1}(\mathcal{X}_t) - F^l_t(\mathcal{X}_t) \right\|_2^2,
\end{equation}
where $L$ denotes the selected intermediate stages in the encoder, and $C_l, H_l, W_l$ are the dimensions of the $l$-th feature map. By explicitly penalizing representational drift at multiple scales, $\mathcal{L}_{MFM}$ acts as a spatial regularizer that directly supports the feature propagation stability of the S$^3$U.

The overall training objective combines the task-specific supervision with this dual-level distillation strategy:
\begin{equation}
    \mathcal{L}_{total} = \mathcal{L}_{CE} + \lambda_{CBD} \mathbb{D}_{AB}^{(\alpha,\beta)}(p \parallel q_\theta) + \lambda_{MFM} \mathcal{L}_{MFM},
\end{equation}
where $\mathcal{L}_{CE}$ is the standard cross-entropy loss evaluated against the strictly binary label space, and $\lambda_{CBD}$ and $\lambda_{MFM}$ are hyperparameters balancing the distillation terms.

\section{Experiments}

\subsection{Datasets}
To evaluate the DICD performance, we utilize three distinct datasets:
\begin{itemize}
    \item \textbf{SYSU-CD}: Contains 0.5m resolution aerial images. We utilize a subset of 4,000 pairs, partitioned into 3,000 pairs for training and 1,000 pairs for testing.
    \item \textbf{CDD}: Comprises season-varying images from Google Earth. We selected 10,000 pairs of images for training and 3,000 pairs for testing.
    \item \textbf{PRCV}: Comprises $512 \times 512$ multitemporal images focusing on ground object changes. We leverage 6,000 pairs for training and 3,000 pairs for testing.
\end{itemize}

\subsection{Implementation Details}

Models are all implemented in PyTorch and trained on a single RTX 4090D GPU for 50 epochs. The network is optimized via Adam with initial learning rate $5 \times 10^{-4}$. Distillation weights are empirically set to $\lambda_c = 0.1$ and $\lambda_{fm} = 0.1$ to balance knowledge retention and acquisition. For ablation studies, $\alpha = 0.8$, $\beta = 0.8$, and temperature $T = 1.0$ are fixed.

Evaluation metrics include F1-score and IoU, with all results reported in percentages. To explicitly evaluate the model's resistance to catastrophic forgetting during incremental learning sequences, we introduce the memory retention metrics, Mem F1 and Mem IoU. Specifically, the retention is calculated as the ratio of the model's performance on the old domain after incremental learning ($\mathcal{M}_{new}$) to its original baseline performance on that domain ($\mathcal{M}_{base}$):

\begin{equation}
    Mem \mathcal{M} = \frac{\mathcal{M}_{new}}{\mathcal{M}_{base}} \times 100\%
\end{equation}

where $\mathcal{M}$ denotes either the F1-score or IoU. Multi-stage performance is denoted as X$\rightarrow$Y or X$\rightarrow$Y$\rightarrow$Z.

\subsection{Performance Comparison}
Our framework is compared with static models including CNNs, Transformers, Mambas and incremental learning (IL) baselines to evaluate catastrophic forgetting mitigation under a binary label space.

\FloatBarrier  % 强制浮动体在此处排版，避免表格飘到下一页
\begin{table*}
\centering
\caption{Accuracy Assessment for Two-Stage Incremental Change Detection. In this and other tables, C indicates CNN-based models, T indicates Transformer-based models, M indicates Mamba-based models, and IL indicates incremental learning baselines. We highlight the highest values in red and the second-highest results in blue.}
\label{tab:two_stage}
\renewcommand{\arraystretch}{1.15} % 微调行高，让表格显得更宽敞
\setlength{\tabcolsep}{1.0mm}      % 进一步缩小列间距（从1.5mm调至1.0mm）
\resizebox{\textwidth}{!}{
% 将固定列宽从 1.4cm 缩小至 1.15cm。内部宽度减小后，\resizebox 整体缩放的比例就会变大，从而实现字体放大。
\newcommand{\fw}[1]{\makebox[1.15cm][c]{#1}}
\begin{tabular}{cccccccccccc}
\toprule
\multirow{3}{*}{Type} & \multirow{3}{*}{Method} & \multicolumn{2}{c}{Base Memory} & \multicolumn{4}{c}{Old Memory} & \multicolumn{4}{c}{New Memory} \\
\cmidrule(lr){3-4} \cmidrule(lr){5-8} \cmidrule(lr){9-12} % 使用 (lr) 修饰符让横线左右两侧略微缩进，形成视觉分组
& & \multicolumn{2}{c}{SYSU} & \multicolumn{2}{c}{SYSU (+C)} & \multicolumn{2}{c}{SYSU (+P)} & \multicolumn{2}{c}{CDD} & \multicolumn{2}{c}{PRCV} \\
\cmidrule(lr){3-4} \cmidrule(lr){5-6} \cmidrule(lr){7-8} \cmidrule(lr){9-10} \cmidrule(lr){11-12}
& & \fw{F1} & \fw{IoU} & \fw{Mem F1} & \fw{Mem IoU} & \fw{Mem F1} & \fw{Mem IoU} & \fw{F1} & \fw{IoU} & \fw{F1} & \fw{IoU} \\
\midrule
\multirow{5}{*}{C} 
& FC-Siam-conc \cite{Daudt2018Siamese}    & 74.86 & 59.82 & 44.95 & 33.81 & 39.05 & 28.62 & 75.16 & 60.28 & 72.14 & 56.42 \\
& FC-Siam-diff \cite{Daudt2018Siamese}    & 75.99 & 61.28 & 43.24 & 32.08 & 40.74 & 29.89 & 77.15 & 62.73 & 71.09 & 55.14 \\
& CDNet \cite{Alcantarilla2018Street}     & 72.87 & 57.32 & 37.92 & 27.97 & 34.33 & 24.94 & 71.65 & 55.79 & 67.76 & 51.24 \\
& LUNet \cite{Papadomanolaki2021Multitask}        & 75.48 & 60.61 & 39.22 & 29.32 & 37.73 & 27.39 & 80.75 & 67.73 & 76.50 & 61.97 \\
& HANet \cite{Han2023HANet}           & 77.46 & 63.21 & 42.33 & 31.45 & 40.56 & 29.48 & 79.67 & 66.22 & 75.33 & 60.42 \\
\midrule
\multirow{4}{*}{T} 
& ChangFormerV1 \cite{Bandara2022Transformer}  & 78.26 & 64.28 & 31.45 & 22.16 & 35.99 & 25.50 & 83.17 & 71.29 & 74.39 & 59.23 \\
& ChangFormerV2 \cite{Bandara2022Transformer}  & 76.47 & 61.90 & 21.87 & 15.19 & 29.45 & 20.52 & 83.38 & 71.65 & 76.61 & 62.10 \\
& ChangFormerV3 \cite{Bandara2022Transformer}  & 79.18 & 65.53 & 26.15 & 18.27 & 34.38 & 24.04 & 80.06 & 66.77 & 74.64 & 59.55 \\
& TransWCD-pixel \cite{Zhao2025TransWCD} & 80.04 & 66.72 & 43.12 & 31.26 & 39.65 & 28.27 & \textcolor{red}{84.72} & \textcolor{red}{73.96} & \textcolor{blue}{78.80} & \textcolor{blue}{64.99} \\
\midrule
\multirow{2}{*}{M} 
& ChangeMamba \cite{Chen2024ChangeMamba}    & \textcolor{blue}{80.58} & \textcolor{blue}{67.48} & 46.75 & 34.39 & 43.11 & 30.57 & 78.31 & 64.33 & 74.29 & 59.10 \\
& RSMamba \cite{Zhao2024RSMamba}        & 80.26 & 67.02 & 45.59 & 33.41 & 38.50 & 27.26 & 81.64 & 69.21 & 73.18 & 57.70 \\
\midrule
\multirow{4}{*}{IL} 
& MDINet (FT) \cite{Weng2024MDINet}    & \multirow{3}{*}{76.43} & \multirow{3}{*}{61.85} & 61.58 & 49.76 & 54.09 & 42.13 & 82.21 & 69.80 & 74.03 & 58.77 \\
& MDINet (FE) \cite{Weng2024MDINet}    &      &      & 52.24 & 40.33 & 46.72 & 35.14 & 79.60 & 66.10 & 71.52 & 55.67 \\
& MDINet (KD) \cite{Weng2024MDINet}    &      &      & \textcolor{blue}{63.29} & \textcolor{blue}{51.58} & \textcolor{blue}{55.70} & \textcolor{blue}{43.72} & 83.03 & 71.03 & 75.09 & 60.11 \\
& Ours            & \textcolor{red}{81.56} & \textcolor{red}{68.85} & \textcolor{red}{70.21} & \textcolor{red}{58.26} & \textcolor{red}{59.45} & \textcolor{red}{46.47} & \textcolor{blue}{83.67} & \textcolor{blue}{71.93} & \textcolor{red}{79.34} & \textcolor{red}{65.75} \\
\bottomrule
\end{tabular}
}
\vspace{-6pt}  % 微调表格下方间距，让文字更紧凑
\end{table*}
\FloatBarrier  % 关闭浮动体屏障，恢复正常排版

\subsubsection{Two-Stage Domain Incremental Change Detection.}

As shown in Table \ref{tab:two_stage}, static architectures prioritize plasticity at the cost of severe catastrophic forgetting. For example, after adapting to CDD, TransWCD-pixel retains only 43.12\% of its base F1 score on SYSU. While incremental baselines like MDINet mitigate this degradation, they still struggle to fully preserve historical knowledge.

In contrast, our proposed DSINet achieves a state-of-the-art balance between stability and plasticity. It delivers the highest knowledge retention across all baselines without compromising its adaptability to new domains, achieving highly competitive F1 scores of 83.67\% on CDD and 79.34\% on PRCV. This optimal tradeoff demonstrates that our dual-selective mechanism effectively disentangles and preserves domain-agnostic structures during new environmental adaptation.

\subsubsection{Three-Stage Domain Incremental Change Detection.}
As the incremental sequence lengthens, accumulated domain shifts severely exacerbate knowledge confusion. Table \ref{tab:three_stage} details the memory retention metrics across the S$\rightarrow$C$\rightarrow$P and S$\rightarrow$P$\rightarrow$C sequences, explicitly tracking the retention of both the base and intermediate domains after adapting to the final target.

\FloatBarrier  % 强制浮动体在此处排版，避免表格飘到下一页
\begin{table*}[htbp]
\centering
\caption{Accuracy Assessment for Three-Stage Incremental Change Detection.}
\label{tab:three_stage}
\renewcommand{\arraystretch}{1.15} % 微调行高，让表格显得更宽敞
\setlength{\tabcolsep}{2mm}      % 调整列间距
\resizebox{\textwidth}{!}{
% 定义固定宽度命令 \fw，保证 8 个数据列的宽度绝对一致
% 因为总列数比上一个表少，这里将宽度设为 1.5cm，以保证缩放后的字体大小与上表接近
\newcommand{\fw}[1]{\makebox[1.5cm][c]{#1}}
\begin{tabular}{cccccccccc}
\toprule
\multirow{3}{*}{Type} & \multirow{3}{*}{Method} & \multicolumn{4}{c}{S$\rightarrow$C$\rightarrow$P Sequence} & \multicolumn{4}{c}{S$\rightarrow$P$\rightarrow$C Sequence} \\
\cmidrule(lr){3-6} \cmidrule(lr){7-10}
 & & \multicolumn{2}{c}{Mem (S)} & \multicolumn{2}{c}{Mem (C)} & \multicolumn{2}{c}{Mem (S)} & \multicolumn{2}{c}{Mem (P)} \\
\cmidrule(lr){3-4} \cmidrule(lr){5-6} \cmidrule(lr){7-8} \cmidrule(lr){9-10}
 & & \fw{Mem F1} & \fw{Mem IoU} & \fw{Mem F1} & \fw{Mem IoU} & \fw{Mem F1} & \fw{Mem IoU} & \fw{Mem F1} & \fw{Mem IoU} \\
\midrule
\multirow{5}{*}{C} 
 & FC-Siam-conc \cite{Daudt2018Siamese}   & 29.46 & 20.51 & 46.41 & 34.24 & 29.13 & 20.47 & 49.31 & 36.28 \\
 & FC-Siam-diff \cite{Daudt2018Siamese}   & 30.78 & 21.63 & 46.68 & 34.76 & 30.57 & 21.64 & 48.94 & 35.80 \\
 & CDNet \cite{Alcantarilla2018Street}          & 27.84 & 19.42 & 40.37 & 30.11 & 23.32 & 15.90 & 43.70 & 31.28 \\
 & LUNet \cite{Papadomanolaki2021Multitask}        & 28.53 & 19.87 & 42.83 & 31.65 & 27.51 & 19.13 & 50.26 & 37.06 \\
 & HANet \cite{Han2023HANet}          & 32.11 & 22.72 & 44.10 & 32.55 & 31.06 & 21.95 & 49.98 & 36.68 \\
\midrule
\multirow{4}{*}{T} 
 & ChangFormerV1 \cite{Bandara2022Transformer}  & 25.78 & 18.11 & 36.37 & 26.47 & 25.28 & 17.64 & 45.36 & 32.94 \\
 & ChangFormerV2 \cite{Bandara2022Transformer}  & 16.93 & 11.32 & 35.12 & 25.52 & 22.43 & 15.52 & 42.69 & 31.16 \\
 & ChangFormerV3 \cite{Bandara2022Transformer}  & 21.37 & 14.31 & 34.71 & 25.63 & 25.63 & 17.29 & 42.81 & 31.08 \\
 & TransWCD-pixel \cite{Zhao2025TransWCD} & 37.85 & 26.93 & 36.43 & 27.20 & 30.42 & 20.87 & 50.96 & 37.78 \\
\midrule
\multirow{2}{*}{M} 
 & ChangeMamba \cite{Chen2024ChangeMamba}    & 39.23 & 28.16 & 45.70 & 33.69 & 40.79 & 29.47 & 52.53 & 38.43 \\
 & RSMamba \cite{Zhao2024RSMamba}        & 33.75 & 23.79 & 43.82 & 32.19 & 35.25 & 24.97 & 46.67 & 33.82 \\
\midrule
\multirow{4}{*}{IL} 
 & MDINet (FT) \cite{Weng2024MDINet}    & 49.60 & 38.05 & \textcolor{blue}{48.26} & \textcolor{blue}{35.83} & 44.89 & 33.91 & \textcolor{blue}{58.69} & \textcolor{blue}{44.68} \\
 & MDINet (FE) \cite{Weng2024MDINet}    & 38.58 & 28.95 & 44.75 & 33.15 & 41.16 & 31.17 & 56.39 & 42.41 \\
 & MDINet (KD) \cite{Weng2024MDINet}    & \textcolor{blue}{50.22} & \textcolor{blue}{38.55} & 47.51 & 34.69 & \textcolor{blue}{45.25} & \textcolor{blue}{34.30} & 57.06 & 42.14 \\
 & Ours           & \textcolor{red}{55.64} & \textcolor{red}{43.59} & \textcolor{red}{50.45} & \textcolor{red}{36.93} & \textcolor{red}{49.21} & \textcolor{red}{37.39} & \textcolor{red}{60.75} & \textcolor{red}{45.58} \\
\bottomrule
\end{tabular}
}
\vspace{-6pt}  % 微调表格下方间距，让文字更紧凑
\end{table*}
\FloatBarrier  % 关闭浮动体屏障，恢复正常排版

As shown in Table \ref{tab:three_stage}, extending the evaluation to three-stage sequences severely exacerbates historical knowledge degradation. Static models suffer extreme performance drops on old domains; after sequentially learning CDD and PRCV, ChangFormerV2 retains an abysmal 16.93\% of its original F1 score on the base SYSU dataset. While incremental baselines like MDINet attempt to mitigate this, their standard KD produces over-smoothed probability distributions that gradually erode sharp spatial boundaries over extended steps.

\begin{figure*}[htbp]
\centering
\includegraphics[width=1.0\textwidth]{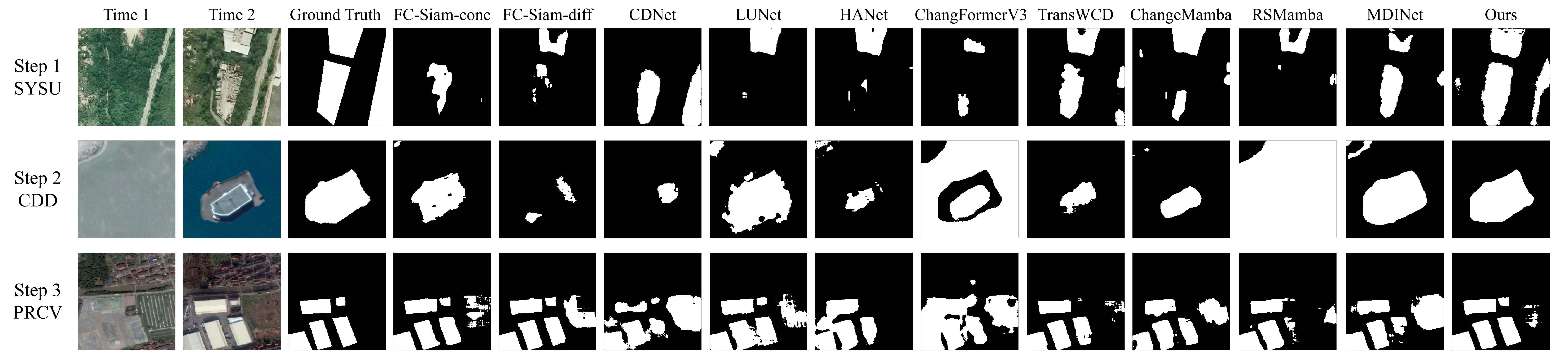} 
\caption{Qualitative comparison on the three-stage incremental sequence (SYSU $\rightarrow$ CDD $\rightarrow$ PRCV). Visualization results on the base domain (SYSU), first new domain (CDD), and second new domain (PRCV) after full incremental training.}
\label{fig:three_stage_visual}
\end{figure*}

Under these challenging conditions, DSINet exhibits superior learning stability and effectively resists catastrophic forgetting. Our framework consistently achieves the highest retention metrics across all historical domains. After completing the full S$\rightarrow$C$\rightarrow$P sequence, DSINet successfully retains 55.64\% Mem F1 on the base SYSU domain and 50.45\% Mem F1 on the intermediate CDD domain, outperforming the best MDINet variants by a significant margin. This quantitative advantage is visually corroborated in Figure \ref{fig:three_stage_visual}, demonstrating that DSINet maintains sharp spatial boundaries and accurate change masks across all stages without the structural erosion typical of other baselines.

\subsection{Ablation Study}

To validate the contributions of our proposed components, we conduct an ablation study evaluating the S$^3$U and CBD. Table \ref{tab:ablation} presents the mean metrics across both two-stage and three-stage incremental sequences. Metrics represent the mean performance across all seen domains.

\begin{table*}[htbp]
\centering
\caption{Ablation study of the proposed components (S$^3$U and CBD) on both two-stage and three-stage incremental sequences.}
\label{tab:ablation}
\renewcommand{\arraystretch}{1.15}
\setlength{\tabcolsep}{2mm} % 减小列间距，配合固定列宽使用
\resizebox{\textwidth}{!}{
% 定义固定列宽，确保所有指标列视觉上完全对齐
\newcommand{\fw}[1]{\makebox[1.6cm][c]{#1}}
\begin{tabular}{lcccccccc}
\toprule
\multirow{2}{*}{Method} & \multicolumn{2}{c}{S$\rightarrow$C} & \multicolumn{2}{c}{S$\rightarrow$P} & \multicolumn{2}{c}{S$\rightarrow$C$\rightarrow$P} & \multicolumn{2}{c}{S$\rightarrow$P$\rightarrow$C} \\
\cmidrule(lr){2-3} \cmidrule(lr){4-5} \cmidrule(lr){6-7} \cmidrule(lr){8-9}
 & \fw{mF1} & \fw{mIoU} & \fw{mF1} & \fw{mIoU} & \fw{mF1} & \fw{mIoU} & \fw{mF1} & \fw{mIoU} \\
\midrule
Baseline & 65.71 & 48.93 & 61.73 & 44.64 & 55.94 & 38.83 & 55.14 & 38.07 \\
+ S$^3$U & 67.31 & 50.73 & 63.25 & 46.25 & 58.40 & 41.24 & 59.82 & 42.67 \\
+ CBD    & 68.42 & 52.01 & 65.79 & 49.02 & 56.35 & 39.23 & 57.26 & 40.11 \\
\midrule
\textbf{+ S$^3$U + CBD (Ours)} & \textbf{70.77} & \textbf{54.76} & \textbf{69.13} & \textbf{52.82} & \textbf{59.83} & \textbf{42.68} & \textbf{60.90} & \textbf{43.78} \\
\bottomrule
\end{tabular}
}
\end{table*}

\textbf{Effectiveness of S$^3$U.} Integrating the S$^3$U into the baseline consistently improves retention metrics across all sequences. In the challenging S$\rightarrow$P$\rightarrow$C sequence, S$^3$U elevates the baseline mF1 from 55.14\% to 59.82\%, demonstrating its ability to effectively filter domain-specific noise during feature propagation. By preserving domain-agnostic spatial structures, the network inherently avoids feature confusion when adapting to new geographic regions.

\textbf{Effectiveness of CBD.} Applying CBD alone robustly boosts overall performance, as seen in the S$\rightarrow$P sequence where it improves the baseline mF1 from 61.73\% to 65.79\%. This confirms that explicitly balancing hardness and confidence during distillation prevents the over-smoothing of probability mass, allowing the model to maintain sharper structural boundaries over continuous updates compared to standard KD.

\textbf{Synergistic Effect.} The combination of S$^3$U and CBD yields the most significant gains, establishing a strong complementary relationship. While S$^3$U stabilizes internal feature representations, CBD stabilizes external knowledge transfer. The combined model substantially outperforms either component alone by achieving 59.83\% mF1 and 42.68\% mIoU in the S$\rightarrow$C$\rightarrow$P sequence, firmly validating the necessity and effectiveness of our proposed framework for long-sequence DICD.

\section{Conclusion}

In this paper, we proposed the DSINet to tackle the severe catastrophic forgetting and representation confusion inherent in DICD. To address the structural mismatch caused by shifting geographic domains under a fixed binary label space, our framework establishes step-wise stability across both feature propagation and knowledge transfer. At the feature level, the S$^3$U leverages the input-dependent selection of SSMs to dynamically preserve domain-agnostic spatial structures while filtering out domain-specific noise. At the optimization level, the CBD strategy mitigates the distribution matching dilemma of standard knowledge distillation. By explicitly balancing hardness and confidence concentration, CBD prevents over-smoothing and mode collapse, ensuring reliable probability mass allocation over continuous incremental updates.

Extensive experiments on long domain sequences validate that DSINet successfully maintains high retention of historical knowledge, preserving sharp spatial boundaries without compromising its ability to adapt to new environments. Ultimately, DSINet provides a highly scalable and effective solution for continuous earth observation, paving the way for more resilient incremental learning systems in practical remote sensing applications.

\nocite{*}
%
% ---- Bibliography ----
%
% 1. 指定样式文件 splncs04.bst
\bibliographystyle{splncs04}

% 2. 指定你的 .bib 文件名（假设文件名是 rf.bib，这里只写 rf）
\bibliography{rf}

@article{Cheng2024Review,
  author    = {Cheng, Guangliang and Huang, Yunmeng and Li, Xiangtai and Lyu, Shuchang and Xu, Zhaoyang and Zhao, Hongbo and Zhao, Qi and Xiang, Shiming},
  title     = {Change Detection Methods for Remote Sensing in the Last Decade: A Comprehensive Review},
  journal   = {Remote Sensing},
  volume    = {16},
  number    = {13},
  pages     = {2355},
  year      = {2024},
  doi       = {10.3390/rs16132355}
}

@article{vandeVen2022Three,
  author    = {van de Ven, Gido M. and Tuytelaars, Tinne and Tolias, Andreas Savas},
  title     = {Three types of incremental learning},
  journal   = {Nature Machine Intelligence},
  volume    = {4},
  pages     = {1185--1197},
  year      = {2022},
  doi       = {10.1038/s42256-022-00568-3}
}

@article{Huang2024Domain,
  author    = {Huang, Wubiao and Ding, Mingtao and Deng, Fei},
  title     = {Domain-Incremental Learning for Remote Sensing Semantic Segmentation With Multifeature Constraints in Graph Space},
  journal   = {IEEE Transactions on Geoscience and Remote Sensing},
  volume    = {62},
  pages     = {1--15},
  year      = {2024},
  doi       = {10.1109/TGRS.2024.3481875}
}

@article{Wang2023Survey,
  author    = {Wang, Liyuan and Zhang, Xingxing and Su, Hang and Zhu, Jun},
  title     = {A Comprehensive Survey of Continual Learning: Theory, Method and Application},
  journal   = {IEEE Transactions on Pattern Analysis and Machine Intelligence},
  volume    = {46},
  pages     = {5362--5383},
  year      = {2023},
  doi       = {10.1109/TPAMI.2024.3367329}
}

@article{Lee2025ERPASS,
  author    = {Lee, Yeseok and Lee, Donghyeon and Kwak, Taehong and Kim, Yongil},
  title     = {ER-PASS: Experience Replay with Performance-Aware Submodular Sampling for Domain-Incremental Learning in Remote Sensing},
  journal   = {Preprints},
  year      = {2025},
  doi       = {10.3390/rs17183233}
}

@inproceedings{Daudt2018Siamese,
  author    = {Caye Daudt, Rodrigo and Le Saux, Bertrand and Boulch, Alexandre},
  title     = {Fully Convolutional Siamese Networks for Change Detection},
  booktitle = {2018 25th IEEE International Conference on Image Processing (ICIP)},
  pages     = {4063--4067},
  year      = {2018},
  doi       = {10.1109/ICIP.2018.8451652}
}

@article{Han2023HANet,
  author    = {Han, Chengxi and Wu, Chen and Guo, Haonan and Hu, Meiqi and Chen, Hongruixuan},
  title     = {HANet: A Hierarchical Attention Network for Change Detection With Bitemporal Very-High-Resolution Remote Sensing Images},
  journal   = {IEEE Journal of Selected Topics in Applied Earth Observations and Remote Sensing},
  volume    = {16},
  pages     = {3867--3878},
  year      = {2023},
  doi       = {10.1109/JSTARS.2023.3264802}
}

@inproceedings{Bandara2022Transformer,
  author    = {Bandara, Wele Gedara Chaminda and Patel, Vishal M.},
  title     = {A Transformer-Based Siamese Network for Change Detection},
  booktitle = {IGARSS 2022 - 2022 IEEE International Geoscience and Remote Sensing Symposium},
  pages     = {207--210},
  year      = {2022},
  doi       = {10.1109/IGARSS46834.2022.9883686}
}

@article{Alcantarilla2018Street,
  author    = {Alcantarilla, Pablo F. and Stent, Simon and Ros, Germ{\'a}n and Arroyo, Roberto and Gherardi, Riccardo},
  title     = {Street-view change detection with deconvolutional networks},
  journal   = {Autonomous Robots},
  volume    = {42},
  number    = {7},
  pages     = {1301--1322},
  year      = {2018},
  doi       = {10.1007/s10514-018-9734-5}
}

@article{Papadomanolaki2021Multitask,
  author    = {Papadomanolaki, Maria and Vakalopoulou, Maria and Karantzalos, Konstantinos},
  title     = {A Deep Multitask Learning Framework Coupling Semantic Segmentation and Fully Convolutional LSTM Networks for Urban Change Detection},
  journal   = {IEEE Transactions on Geoscience and Remote Sensing},
  volume    = {59},
  number    = {9},
  pages     = {7651--7668},
  year      = {2021},
  doi       = {10.1109/TGRS.2021.3055584}
}

@article{Zhao2025TransWCD,
  author    = {Zhao, Zhenghui and Ru, Lixiang and Wu, Chen and Wang, Di},
  title     = {TransWCD: Scene-Adaptive Joint Constrained Framework for Weakly Supervised Change Detection},
  journal   = {IEEE Transactions on Geoscience and Remote Sensing},
  volume    = {63},
  pages     = {1--12},
  year      = {2025},
  doi       = {10.1109/TGRS.2025.3545051}
}

@article{Zhao2024RSMamba,
  author    = {Zhao, Sijie and Chen, Hao and Zhang, Xue-liang and Xiao, Pengfeng and Lei, Bai and Wanli, Ouyang},
  title     = {RS-Mamba for Large Remote Sensing Image Dense Prediction},
  journal   = {IEEE Transactions on Geoscience and Remote Sensing},
  volume    = {62},
  pages     = {1--14},
  year      = {2024},
  doi       = {10.1109/TGRS.2024.3425540}
}

@article{Chen2024ChangeMamba,
  author    = {Chen, Hongruixuan and Song, Jian and Han, Chengxi and Xia, Junshi and Yokoya, Naoto},
  title     = {ChangeMamba: Remote Sensing Change Detection With Spatiotemporal State Space Model},
  journal   = {IEEE Transactions on Geoscience and Remote Sensing},
  volume    = {62},
  pages     = {1--20},
  year      = {2024},
  doi       = {10.1109/TGRS.2024.3417253}
}

@article{Weng2024MDINet,
  author    = {Weng, Lean and Yang, Wenqing and Hu, Boni and Han, Pengcheng and Xue, Shaocheng and Zhang, Yu and Li, Haowei and Jin, Jie and Bu, Shuhui},
  title     = {MDINet: Multidomain Incremental Network for Change Detection},
  journal   = {IEEE Transactions on Geoscience and Remote Sensing},
  volume    = {62},
  pages     = {1--15},
  year      = {2024},
  doi       = {10.1109/TGRS.2023.3348878}
}

@inproceedings{Wang2025ABKD,
  author    = {Wang, Guanghui and Yang, Zhiyong and Wang, Zitai and Wang, Shi and Xu, Qianqian and Huang, Qingming},
  title     = {{ABKD}: Pursuing a Proper Allocation of the Probability Mass in Knowledge Distillation via $\alpha$-$\beta$-Divergence},
  booktitle = {Proceedings of the 42nd International Conference on Machine Learning (ICML)},
  year      = {2025}
}

@article{Himeur2025Survey,
  author    = {Himeur, Yassine and Aburaed, Nour and Elharrouss, Omar and Varlamis, Iraklis and Atalla, Shadi and Mansoor, Wathiq and Al-Ahmad, Hussain},
  title     = {Applications of knowledge distillation in remote sensing: A survey},
  journal   = {Information Fusion},
  volume    = {115},
  pages     = {102742},
  year      = {2025},
  doi       = {10.1016/j.inffus.2024.102742}
}

\end{document}